\documentclass[journal]{IEEEtran}
\usepackage{tikz}
\usepackage{amsmath,amsfonts}
\usepackage{array}
\usepackage[caption=false,font=normalsize,labelfont=sf,textfont=sf]{subfig}
\usepackage{textcomp}
\usepackage{stfloats}
\usepackage{url}
\usepackage{verbatim}
\usepackage{graphicx}
\usepackage{cite}
\usepackage{xcolor}

\usepackage[linesnumbered,ruled,vlined,algo2e]{algorithm2e}

\newif\ifhighlight \highlightfalse
\definecolor{revcolor}{RGB}{0,70,200}
\newcommand{\rev}[1]{\ifhighlight{\color{revcolor}#1}\else#1\fi}
\newcommand{\newrefkeys}{Isele2018,Aljundi2019GSS,Aljundi2019MIR,Aljundi2019TaskFree,Chaudhry2019,Vitter1985,Orendurff2008,MoulaeeConradsson2025,Pellegrini2020,Hayes2020,Yuan2024,schaul2016prioritized}
\makeatletter
\@for\@nrk:=\newrefkeys\do{\expandafter\let\csname newref@\@nrk\endcsname\@empty}
\def\rev@bibcolor#1{\ifhighlight\ifcsname newref@#1\endcsname\color{revcolor}\else\normalcolor\fi\fi}
\let\rev@origthebibliography\thebibliography
\def\thebibliography#1{\rev@origthebibliography{#1}\let\rev@origbibitem\bibitem
  \let\rev@origbiblabel\@biblabel\def\@biblabel##1{{\rev@bibcolor\rev@curkey\rev@origbiblabel{##1}}}%
  \def\bibitem{\@ifnextchar[\rev@lbibitem\rev@bibitem}}
\def\rev@curkey{}
\def\rev@bibitem#1{\def\rev@curkey{#1}\rev@origbibitem{#1}\rev@bibcolor{#1}}
\def\rev@lbibitem[#1]#2{\def\rev@curkey{#2}\rev@origbibitem[#1]{#2}\rev@bibcolor{#2}}
\makeatother
\ifdefined\anonymousversion \newcommand{\anon}[2]{#2}\else \newcommand{\anon}[2]{#1}\fi

\begin{document}

\title{{\fontsize{22pt}{30pt}\selectfont Continual Online Personalization of Exoskeleton Control via \rev{Gait Feature-Aware} Experience Replay}}

\author{\anon{Changseob Song and Inseung Kang,~\IEEEmembership{Member,~IEEE}}{Anonymous Authors}%

\thanks{This work was supported in part by the \anon{NIH R21 Award 1R21EB037268-01}{xxxx} and in part by the \anon{Kwanjeong Educational Foundation}{xxxx}.
Corresponding author: \anon{Changseob Song}{Anonymous Author}, {\tt\footnotesize \anon{changseob@cmu.edu}{xxxx@xxxx}}.

This work involved human subjects or animals in its research.
Approval of all ethical and experimental procedures and protocols was granted by the \anon{Carnegie Mellon University}{xxxx} Institutional Review Board.

\anon{C. Song and I. Kang are with the Department of Mechanical Engineering, Carnegie Mellon University, Pittsburgh, PA 15213, USA.}{Anonymous Authors are with the Department of xxxx xxxx, xxxx, xxxx, xxxx, xxxx}}}

\maketitle

\begin{abstract}
Personalizing exoskeleton control remains a critical challenge for clinical users with gait disabilities.
Online adaptation (OA) addresses this by adapting \rev{the gait phase estimator} in real time to user variability, device fit, and locomotor tasks.
However, OA learns from a continual data stream and can catastrophically forget previously learned locomotor contexts.
Here, we develop a \rev{gait feature-aware} experience replay framework for OA of \rev{a hip exoskeleton gait phase estimator}.
\rev{The replay buffer is organized in a gait feature space}, so replay samples are selected without explicit task labels.
We evaluated the framework \rev{with able-bodied participants emulating} hemiplegic gait across multiple forgetting scenarios with speed and incline transitions.
Compared to OA without replay, which exhibited catastrophic forgetting during task transitions, our framework \rev{reduced gait phase root mean square error (RMSE) by 39.2\% and torque command RMSE by 28.6\%}.
\rev{Compared with matched-budget reservoir baselines, it kept more samples of the non-ongoing tasks available for replay during prolonged single-task walking, although all replay methods achieved comparable errors in this study.}
This demonstrates that our framework \rev{maintains real-time personalization of gait phase estimation} across diverse locomotor contexts\rev{, although validation in clinical populations during daily ambulation remains future work}.
\end{abstract}


\section{Introduction}

Robotic exoskeletons have the potential to support community ambulation in daily life \cite{sawicki2020, Gao2025, siviy2023}.
Beyond improving the walking economy of able-bodied users \cite{sawicki2020}, exoskeletons have demonstrated benefits for clinical populations with motor impairments by improving walking economy \cite{Pruyn2026, Kang2025, Gunnell2025, Awad2017}, gait stability \cite{Awad2017, Archangeli2026, Lerner2017SciTranslMed, Kim2024}, pain \cite{Divekar2025, Zhang2025}, and user preference \cite{Ingraham2022, Lee2023}.
While several studies have developed exoskeleton controllers that improve gait outcomes for clinical users, these controllers rely on manual tuning of control parameters \cite{Archangeli2026, Pruyn2026,  Gunnell2025} and extensive optimization time for each target user \cite{Awad2017}.
Personalizing exoskeleton control rapidly, within a short amount of user experience, and efficiently, without a labor-intensive tuning process, has therefore become an important agenda in the wearable robotics field.
This challenge intensifies for clinical populations such as stroke survivors, whose substantial inter-subject gait variability makes effective assistance particularly difficult to achieve \cite{Winter1984, Olney1996, Balaban2014}.

Recent advances in the wearable robotics domain have been largely driven by machine learning (ML)-based control methods \cite{Luo2024, Park2026, Molinaro2024, Molinaro2024SciRob, Lee2024, Kang2025, Kang2021, Kang2019}.
Trained on large datasets spanning diverse locomotor tasks, ML-driven exoskeleton controllers have shown promise for assisting users across various real-world scenarios \cite{Molinaro2024, Kang2021, Park2026, Lee2024}, and can estimate diverse user states in real time, such as biological joint moment \cite{Molinaro2024, Molinaro2024SciRob}, gait phase \cite{Kang2019, Kang2021, Kang2025, Lee2024}, and activity mode \cite{Lee2024, Kang2022, Maldonado2025, Johnson2025}, which can be mapped into exoskeleton torque commands.
However, these data-driven controllers are inherently bounded by the coverage of their training distribution and degrade under out-of-distribution conditions for unseen users and tasks, requiring retraining on downstream clinical users \cite{Kang2025, Maldonado2025, Johnson2025}.
This hinders the applicability of exoskeletons, particularly when users cannot easily access gait labs with expensive instrumentation.

To personalize these static controllers, several studies have proposed online adaptation (OA) strategies that adjust the controller toward target users and tasks in real time \cite{Kang2025, Maldonado2025, Johnson2023, Johnson2025}.
OA has shown promise in accommodating user gait variability, differences in device fit relative to the pre-trained hardware setup, and sensor signal shift induced by torque application.
For instance, Kang \textit{et al.} developed an OA framework that adapts a gait phase estimator toward stroke gait in real time \cite{Kang2025}, and Johnson \textit{et al.} online-adapted walking speed estimators for prosthetic applications \cite{Johnson2023}.
While these studies personalized user state estimators in real time, their implementation was restricted to single-task scenarios.
However, exoskeleton personalization requires the controller to adapt concurrently across a wide range of locomotor tasks, which single-task frameworks cannot provide.

In continual learning scenarios involving multiple tasks, models are susceptible to catastrophic forgetting, in which information from previously learned tasks is lost through overfitting to the current task \cite{Wang2024, Delange2021, Mermillod2013}.
To mitigate this, the ML community has developed diverse solutions, including experience replay, regularization, and architecture-based methods \cite{Wang2024, Delange2021, Mermillod2013}.
Among these, experience replay is one of the most effective approaches, storing previously encountered experiences and replaying them throughout continual adaptation \cite{Rolnick2018}.
\rev{During deployment, where task labels are often unavailable, task-free replay methods \cite{Aljundi2019TaskFree} manage replay by uniform reservoir sampling of the data stream \cite{Vitter1985, Rolnick2018, Chaudhry2019} or by selecting samples based on state-space coverage \cite{Isele2018}, gradient diversity \cite{Aljundi2019GSS}, or interference with the upcoming update \cite{Aljundi2019MIR}.}
\rev{Methods that select samples by their location in a representation space can retain old or rarely performed tasks, but they require class labels \cite{Chaudhry2019} or compare each incoming sample with the stored samples \cite{Isele2018, Aljundi2019GSS}.}

In the wearable robotics field, experience replay has so far been applied to prosthetic studies, including a pipeline that replays bins across different walking speeds \cite{Maldonado2025} and a stability-plasticity balancing pipeline that adjusts the learning rate and the replayed task \cite{Johnson2025}.
While successful within the prosthetics domain, these approaches leave two limitations unaddressed.
First, no exoskeleton study has demonstrated OA across multimodal task contexts, particularly under the sparser sensing modalities available to exoskeletons relative to prosthetic devices.
Second, prior studies \cite{Maldonado2025, Johnson2025} \rev{adapted walking speed estimators rather than estimators of assistance timing, and they} assumed known task labels when organizing the replay bins, an assumption that does not hold in real-world deployment.

In this work, we incorporate \rev{gait feature-aware} experience replay into the OA of \rev{a hip exoskeleton gait phase estimator} \rev{and evaluate it in able-bodied participants walking with emulated hemiplegic gait}, addressing the two limitations above.
The estimated gait phase is mapped into a torque profile to deliver assistance. Because gait phase is a universal high-level user state, adapting the estimator adjusts the assistance timing \cite{Kang2025, Young2017, Slade2022personalizing}.
To maintain stable estimation across diverse tasks, the framework continually replays the locomotor tasks the user has previously encountered.
To remove the need for explicit task labeling, it organizes the replay buffer using the encoded features of gait data, allowing replay samples to be selected without locomotor context labels.
With this \rev{gait feature-aware} OA framework, we expect the exoskeleton to provide personalized assistance across diverse locomotor tasks while maintaining stability under continual adaptation.

\section{Method}

\subsection{Robotic Hip Exoskeleton}

\begin{figure}
    \centering
    \includegraphics[width=0.49\textwidth]{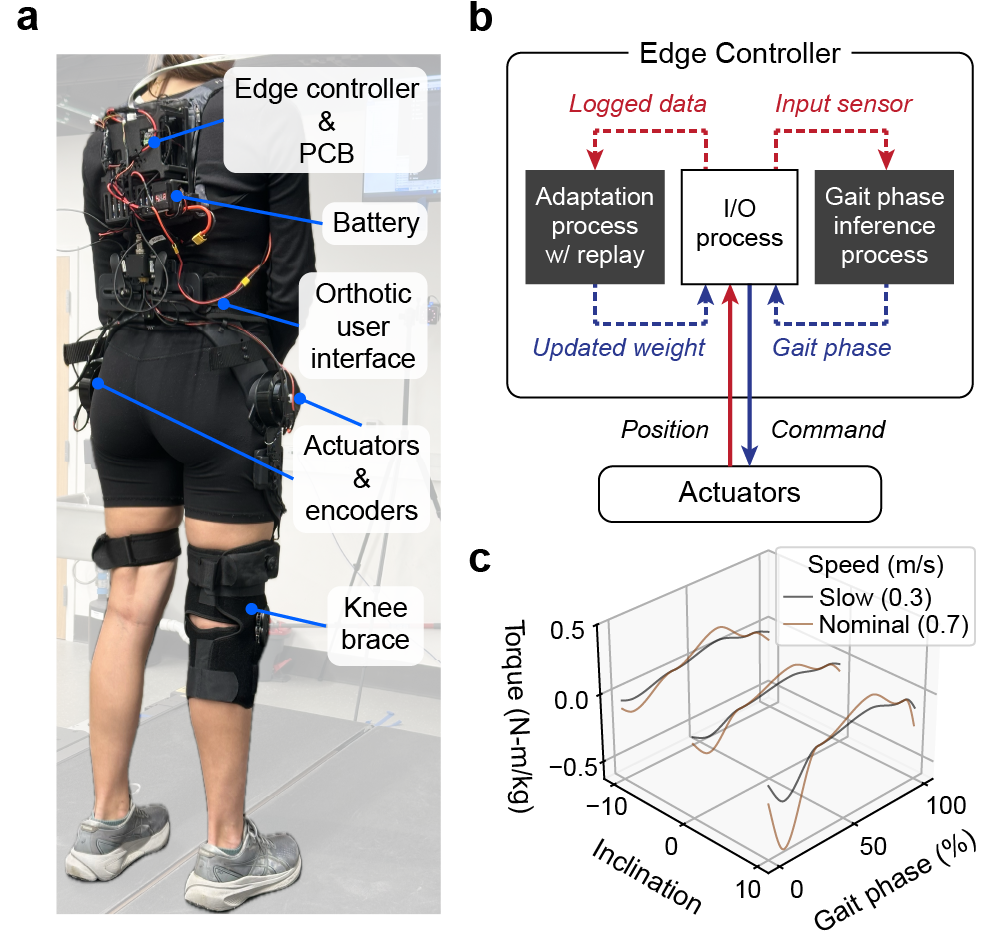}
    \caption{
        Hip exoskeleton hardware and software.
        (a) Robotic hip exoskeleton designed to assist the user’s bilateral hip motion during locomotion.
        (b) Real-time control flow within the edge controller unit.
Dashed lines indicate inter-process communication between parallel software modules, while solid lines show peripheral communication between the controller’s I/O process and the physical actuators.
Black shaded boxes show GPU-accelerated processes, and the white box indicates a CPU-centric process.
        (c) Spline profiles for different walking inclines and speeds.
This mid-level spline profile maps high-level gait phase into low-level torque command.
        }
    \label{exo_hw_sw}
\end{figure}

We developed a robotic hip exoskeleton capable of applying bilateral torque (Fig. \ref{exo_hw_sw}a).
The device had a total mass of 4.5 kg and provided joint torque up to 22 N-m in the sagittal plane, powered by a lithium-polymer battery (HRB-POWER).
Two quasi-direct-drive brushless DC motors (AK80-9, CubeMars) actuated the hip joints, each equipped with an encoder that measured the hip joint kinematics.
The edge controller (Jetson Orin Nano, NVIDIA) ran a CPU-centric I/O process that managed joint kinematics acquisition and torque command transmission via CAN communication at 100 Hz (Fig.~\ref{exo_hw_sw}b).
Operating in parallel, a GPU-accelerated gait phase inference process estimated the user's gait state by retrieving encoder signals through asynchronous inter-process communication.
To ensure deterministic control at 100 Hz, the I/O process utilized a zero-order hold strategy, reusing the previous phase estimate when a GPU update was not available within the control cycle.
Simultaneously, an adaptation process \rev{received an adaptation sample containing the two most recent gait cycles at every third heel strike of each limb} and updated the weights of the final linear layer, which were then queued back to the I/O process.
These three modules operated concurrently within a parallelized software architecture. Before each trial, both GPU processes were warmed up with dummy inputs to absorb one-time initialization delays.

The estimated gait phase, representing the high-level user state, was mapped to torque commands.
\rev{Although the gait phase estimator and its OA, including the replay buffer, did not use task context, the corresponding mid-level control splines were selected using task labels from the predefined experimental protocol (Fig. \ref{exo_hw_sw}c), a dependency that remains to be addressed for a fully task-independent controller.}
We obtained predefined splines for level-ground (LG) and ramp-ascent (RA) tasks from a prior study on minimizing energy expenditure \cite{Franks2022}.
Ramp-descent (RD) splines were extrapolated by scaling the LG splines according to the relationship between LG and RD biological joint moments across eleven able-bodied participants, and reference profiles were linearly scaled across speeds.
The resulting torque commands were sent to the low-level controller, which used a torque-tracking loop embedded in the motor driver of the actuator assembly.

\subsection{Gait Phase Estimation}
We developed a deep learning-based gait phase estimator following previous studies \cite{Kang2019, Kang2021}, which served as the pre-trained estimator and the initial seed for online personalization.
The estimator was a temporal convolutional network (TCN), a widely used architecture in the exoskeleton field for modeling time-series data \cite{Molinaro2024, Molinaro2024SciRob}.
To build the pre-trained estimator, we collected 15 hours of able-bodied data from 11 participants.
The dataset spanned five inclinations of -10, -5, 0, 5, and 10 degrees.
The LG data included seven speeds, from 0.2 to 1.4 m/s in 0.2 m/s increments, along with transient speed conditions, while the RD and RA conditions each included four speeds, from 0.4 to 1.0 m/s in 0.2 m/s increments, with transient conditions.

\begin{figure*}[t]
    \centering
    \includegraphics[width=\textwidth]{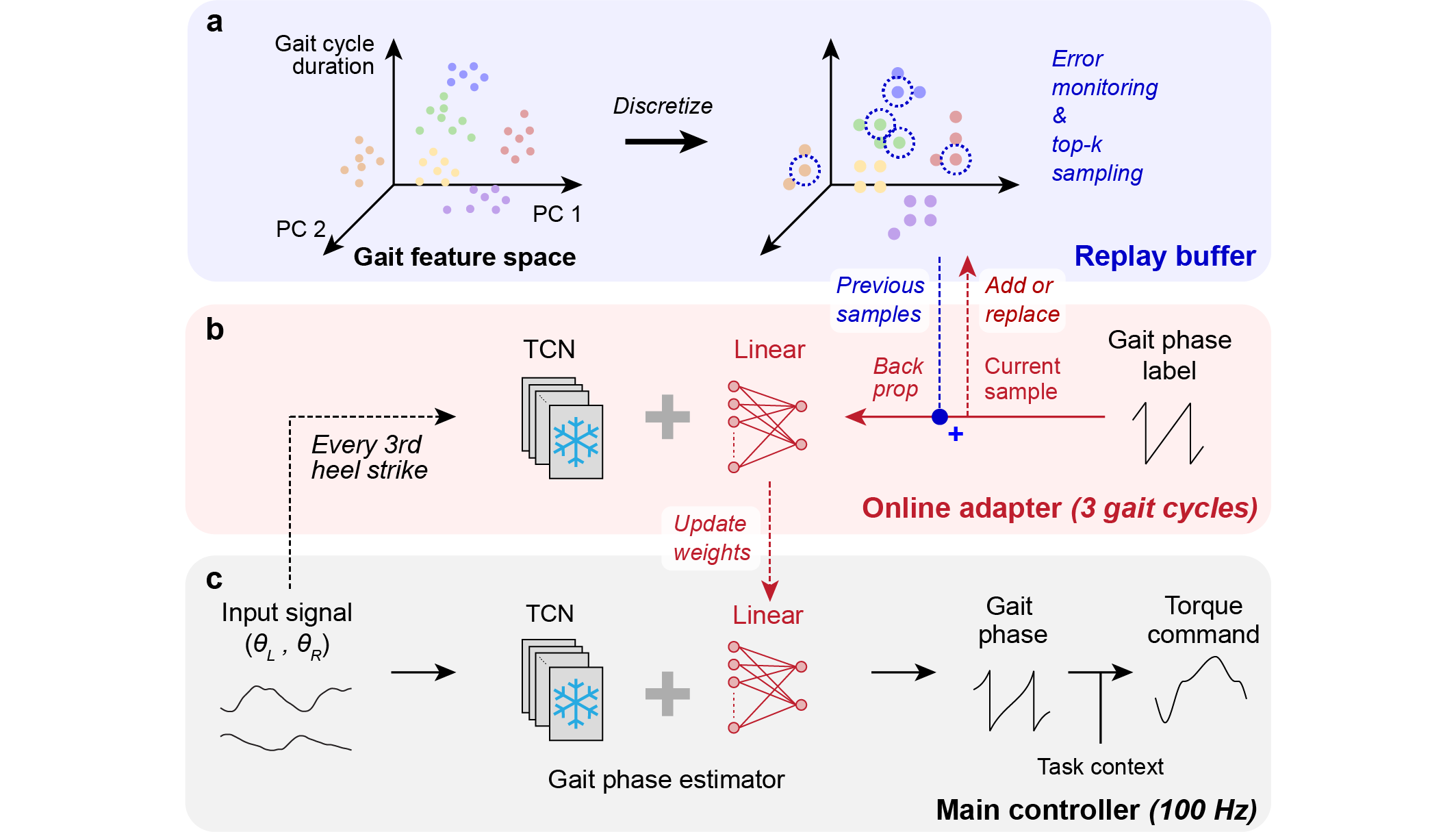}
    \caption{
        \rev{Gait feature-aware} experience replay framework for online adaptation (OA) \rev{of the gait phase estimator}.
The overall architecture includes (a) a replay buffer \rev{organized in the gait feature space}, (b) an OA module that updates the gait phase estimator’s neural network, and (c) a main control loop that receives input signals, predicts the gait phase, and commands torque.
In the main controller, the estimated gait phase is mapped into torque commands using the task context and predefined \rev{torque profiles} in Fig. \ref{exo_hw_sw}c.
The OA module is triggered by the main controller every third heel strike and retrieves data from two gait cycles.
It then generates a gait phase label that is time-synchronized with the streamed input sensor data.
Using this user-specific data, the module updates the weights of the final linear layer of the temporal convolutional network (TCN), while the preceding TCN feature-extracting blocks remain frozen.
The newly personalized weights are subsequently transferred to the main process.
To prevent catastrophic forgetting, the adaptation module utilizes a replay buffer that saves the user’s \rev{gait features}.
As new data arrives, the buffer is updated by adding new data or replacing overlapping \rev{features}.
The adaptation module monitors estimation accuracy across all experienced tasks, prioritizing the top four lowest-accuracy \rev{replay samples} to combine with the current \rev{sample} to form a training dataset.
        }
    \label{OA_framework}
\end{figure*}

The TCN estimated the user's gait phase from bilateral hip encoder signals, forming two input channels (Fig. \ref{OA_framework}c).
We estimated the gait phase of each side separately to simplify the backward labeling of OA and the organization of the replay buffer.
Similar to prior gait phase studies \cite{Kang2019, Kang2021, Kang2025}, we used a Cartesian representation ($x = \cos(\phi)$, $y = \sin(\phi)$), where $\phi$ denotes the gait phase, to account for the inherent discontinuity of the gait phase profile at heel strike, where it transitions from 100\% to 0\%, to improve estimator performance.
The model configuration, optimized via hyperparameter sweeping and leave-one-subject-out validation, used a 1-second temporal window, \rev{five residual TCN blocks with dilation factors of 1, 2, 4, 8, and 16, each containing two dilated causal convolution layers with a kernel size of 5 and 80 channels,} a batch size of 16, and a learning rate of $10^{-6}$.
After training, the TCN blocks were exported in TensorRT format \cite{nvidia_tensorrt} to accelerate inference on the edge controller's GPU, while the parameters of the final linear layer were exported in PyTorch format \cite{PyTorch} to allow for real-time weight updates during the adaptation process.

\begin{algorithm2e}[t]
\SetAlgoLined
\DontPrintSemicolon
\caption{Online Adaptation with \rev{Gait Feature-Aware} Experience Replay}
\label{OA_algorithm}

\KwIn{Models $\pi_{\{L,R\}}$, Params $\theta_{\{L,R\}}$, PCA Matrix $\Phi$, \rev{RMSE band $(\tau_{lo}, \tau_{hi})$}, Grid intervals $\delta_{pc}$, $\delta_t$, Warm-up adaptations $n_{min}$, Max replay number $K$}
\KwOut{Updated parameters $\theta$}

\textbf{Initialize:} Buffer $\mathcal{B} \leftarrow \{ \emptyset \}$, $n_{adapt} \leftarrow 0$ \;

\While{True}{
    $\text{SignalReady}()$ \tcp*[r]{\footnotesize Await next sample}
    $x, \text{side} \leftarrow \text{receive}()$ \tcp*[r]{\footnotesize Get input \& L/R side}

    $\pi, \theta \leftarrow \text{SelectModel}(\text{side})$ \tcp*[r]{\footnotesize Load active model weights}

    $z_{t} \leftarrow \text{dim}_t(x)$ \tcp*[r]{\footnotesize Gait cycle duration}
    $z_{pc} \leftarrow \text{PCA}(x, \Phi)$ \tcp*[r]{\footnotesize PC projection}

    $\kappa \leftarrow \left( \lfloor z_{pc}/\delta_{pc} \rfloor, \lfloor z_{t}/\delta_t \rfloor \right)$ \tcp*[r]{\footnotesize Compute grid key}

    \If{$n_{adapt} \geq n_{min}$}{ \tcp*[r]{\footnotesize Start storage after a minimal number of adaptations}
        \If{$\kappa \notin \mathcal{B}[\text{side}]$}{
            $\mathcal{B}[\text{side}][\kappa] \leftarrow x$ \tcp*[r]{\footnotesize Add sample to new cell}
        }
        \Else{
            $\mathcal{B}[\text{side}][\kappa] \leftarrow x$ \tcp*[r]{\footnotesize Replace the stored sample}
        }
    }

    $\mathcal{E} \leftarrow \{ \text{RMSE}(\pi, b) \mid b \in \mathcal{B}[\text{side}] \}$ \tcp*[r]{\footnotesize Evaluate error on all stored samples}

    $\mathcal{S}_{sort} \leftarrow \text{argsort}(\mathcal{E})$ \tcp*[r]{\footnotesize Rank samples by RMSE (descending)}

    $\mathcal{D}_{replay} \leftarrow \text{Top}_K \left( \{ b \in \mathcal{S}_{sort} \mid \rev{\tau_{lo} < \text{RMSE}(b) < \tau_{hi}} \} \right)$ \tcp*[r]{\footnotesize Keep eligible samples \& select top-$K$}

    $\mathcal{D}_{train} \leftarrow \{x\} \cup \mathcal{D}_{replay}$ \tcp*[r]{\footnotesize Combine current and replay data}

    $\theta \leftarrow \rev{\text{Adam}}(\pi, \mathcal{D}_{train})$ \tcp*[r]{\footnotesize Update weights}

    $n_{adapt} \leftarrow n_{adapt} + 1$ \;
    $\text{send}(\theta)$ \tcp*[r]{\footnotesize Sync with main control loop}
}
\end{algorithm2e}

\subsection{Online Adaptation Framework}
We developed an OA framework that (1) continually streamed the input data logged from the main control loop, (2) backward-labeled the gait phase, and (3) updated the estimator network weights by backpropagating with new data from the current user's state (Fig. \ref{OA_framework}b).
The main control loop triggered the OA process every third heel strike.
Heel strikes were detected from ground reaction force signals streamed by a force plate-instrumented treadmill (FIT5, Bertec) used during the experiment. \rev{Heel strikes from force plate signals were used to backward-label a time-synchronized gait phase with the encoder input signals.}
After the dataset was loaded, the final linear layer of the TCN was updated using the Adam optimizer with a learning rate of $5 \times 10^{-5}$, 50 times larger than the learning rate used to train the gait phase estimator offline, in one shuffled pass with a batch size of 16.
The right and left sides of the estimator were updated separately to simplify the labeling and replay storage processes.

We froze the high-level feature extractor, consisting of \rev{the five TCN blocks}, to preserve the features representing generic human locomotion, and fine-tuned only the final linear layer that maps the latent space to the gait phase.
This strategy is commonly used in transfer learning to preserve general features while adapting downstream task-specific features \cite{Yosinski2014}.
Beyond maintaining the stability of the transfer, it also conserves computing resources and reduces update latency, particularly because the TCN blocks contain 18 times more parameters than the final linear layer in our setup.

\begin{figure}
    \centering
    \includegraphics[width=0.49\textwidth]{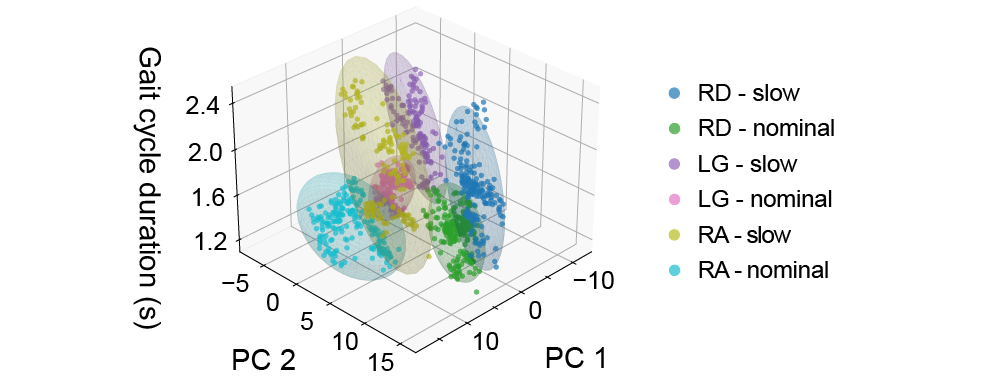}
    \caption{
        \rev{Gait feature space.}
High-dimensional gait data were projected into a 3D \rev{space} comprising the first two principal components, which were derived from able-bodied kinematics data, and the corresponding gait cycle duration.
Each color indicates a distinct locomotor task, and ellipsoids represent the 99\% confidence interval for each cluster.
        }
    \label{gait_manifold}
\end{figure}

\subsection{\rev{PCA-Based Gait Feature}}
We encoded a gait \rev{feature} to distinguish between multimodal locomotor tasks across varying inclines and speeds.
\rev{To replay previous tasks during OA, the replay buffer must distinguish incoming samples from different tasks when storing them.}
While the exoskeleton literature has used discrete task classifiers \cite{Lee2024, Kang2022}, such classifiers themselves degrade under out-of-distribution conditions when encountering new, impaired gait patterns.
We therefore built a feature extractor that projects gait data into a \rev{gait feature space}, preventing redundant replay of similar gait patterns.

To build the feature extractor, we used the same dataset as the pre-trained estimator, consisting of 16,623 gait cycles.
Kinematic data were segmented using reference-side heel strikes and downsampled to 50 points per cycle, and the bilateral signals were concatenated into a 100-dimensional input vector.
We applied principal component analysis (PCA) to these data and retained the first two principal components, which accounted for 77\% of the total variance.
We then used the gait cycle duration of the reference limb as a third axis, establishing a three-dimensional \rev{gait feature space} (Fig. \ref{gait_manifold}).
Validated on the able-bodied dataset used to develop the principal component extractor, the clusters yielded a silhouette score of 0.15 across tasks.
\rev{The projected features overlapped across locomotor tasks but occupied partly different regions}, providing a basis for \rev{feature-aware} replay without explicit task labeling.

\subsection{\rev{Gait Feature-Aware} Replay}
We developed \rev{gait feature-aware} replay to prevent catastrophic forgetting of previous tasks during OA (Algorithm~\ref{OA_algorithm}).
Using the gait \rev{feature} space (Fig.~\ref{gait_manifold}), we \rev{stored samples in a replay grid, which allocated a new cell for each unseen feature region and replaced the stored sample with recent data in a revisited cell}.
When OA was triggered by new gait data, the data were projected into the \rev{feature} space through the principal component transformation.
To avoid maintaining an unbounded number of stored samples, we discretized the projected \rev{space into cells with grid intervals of $\delta_{pc}=2$ (in principal component units) along both principal components and $\delta_t=0.1$~s along gait cycle duration, and each cell held one sample}.
Storage in the replay grid started only after the first four adaptations for each new user so that the estimator could stabilize.
If the cell was unoccupied, the sample was added; otherwise, the sample stored in that cell was replaced with the new data.
After storage, we assessed the current model's performance by tracking the root mean square error (RMSE) of all stored samples against the ground truth, computed on every tenth window of each stored sample, and \rev{only samples with RMSE values between 2.5\% and 20\% were eligible for replay} to ensure efficient and stable replay.
We then \rev{replayed up to four eligible samples} with the highest RMSE, a value optimized by balancing estimation RMSE and update latency (Fig. \ref{effect_of_update_param}b).
Finally, the replay dataset was combined with the current data for backpropagation, and the updated weights were transferred to the main control loop for the gait phase estimator.

\subsection{\rev{Reservoir Baselines}}
\rev{To isolate the contribution of the proposed storage policy from the general benefit of replay, we implemented OA without replay and two reservoir baselines in the same controller (Table~\ref{baselines}).
Both reservoir baselines kept up to $C=50$ samples per limb, close to the largest replay grid in the main experiment, in which the grid of a limb reached at most 55 occupied cells, including the testing trials.
The reservoir baselines were implemented with Algorithm R \cite{Vitter1985}, which appends each incoming sample while the buffer is not full and otherwise lets the $n$-th sample replace a uniformly chosen slot with probability $C/n$.
This yields a uniform random subset of all samples seen so far without task labels, as in reservoir-based experience replay \cite{Rolnick2018, Chaudhry2019}.
At each update, every buffer sample's error was scored in the same way as in the gait feature-aware replay, and only samples with RMSE between 2.5\% and 20\% were eligible.
Reservoir-random drew up to $K=4$ eligible samples uniformly at random, while Reservoir-error selected the $K=4$ eligible samples with the highest RMSE, i.e., the retrieval rule of the gait feature-aware replay.
Random storage and random retrieval used separate pseudo-random number generators, seeded separately for each limb.
A different seed was used for each participant so that each participant's random sequence continued across the protocol.
The comparison also included OA without replay, in which each update trained on the current sample only, as in the main experiment.
The pre-trained model, optimizer, and learning rate were identical across all frameworks, including OA without replay.
The warm-up of four adaptations and the maximum of $K=4$ replayed samples per update were identical across all replay methods.
Comparing Reservoir-random with Reservoir-error isolates the retrieval rule, and comparing Reservoir-error with the gait feature-aware replay isolates the storage policy.}

\begin{table}[t]
\centering
\caption{\rev{Frameworks in the replay comparison experiment}}
\label{baselines}
{\footnotesize\setlength{\tabcolsep}{4pt}
\rev{\begin{tabular}{l l l}
\hline
Framework & Storage (buffer) & Retrieval (up to $K=4$) \\
\hline
Adapted w/o replay & none & none \\
Reservoir-random & reservoir, budget$=50$/limb & random, eligible \\
Reservoir-error & reservoir, budget$=50$/limb & highest RMSE, eligible \\
Gait feature-aware & one sample per grid cell & highest RMSE, eligible \\
\hline
\multicolumn{3}{l}{Eligible: sample RMSE of 2.5--20\%.}
\end{tabular}}}
\end{table}

\begin{figure}
    \centering
    \includegraphics[width=0.49\textwidth]{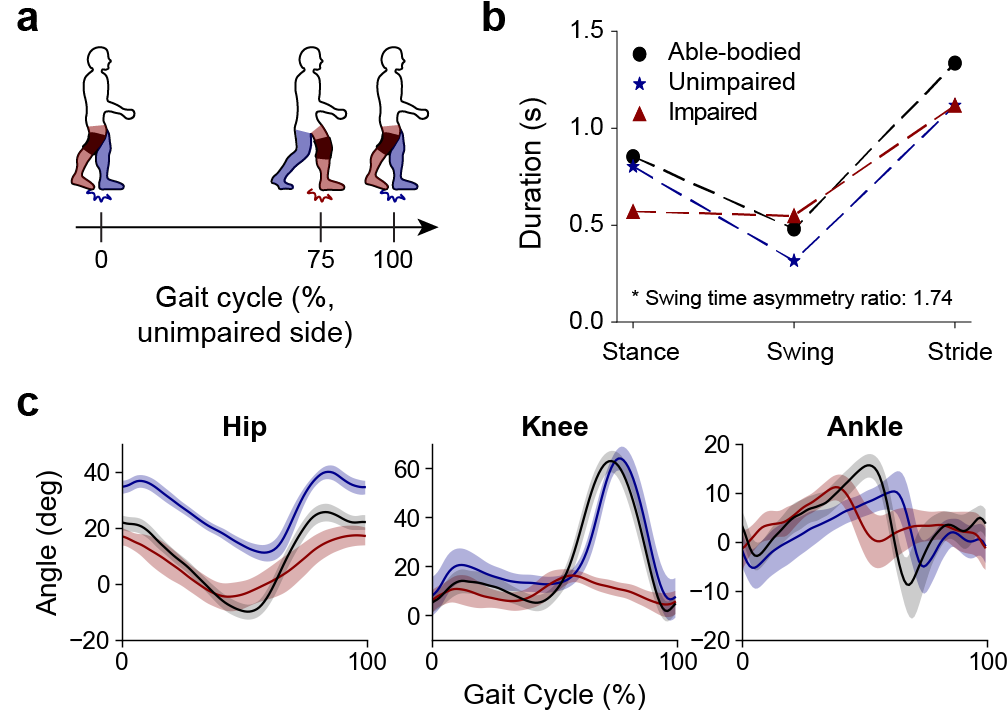}
    \caption{
        Emulated hemiplegic gait.
        (a) Gait cycle timing of impaired- and unimpaired-side heel strikes.
        (b) Stance, swing, and stride durations of emulated gait.
        (c) Sagittal plane joint kinematics of the emulated gait (mean $\pm$ SD, $n=3$) compared to able-bodied gait.
        }
    \label{hemiplegic_gait_simulation}
\end{figure}

\subsection{Hemiplegic Gait Emulation}
We focused on stroke survivors as the target population because of their high inter- and intra-subject gait variability and gait asymmetry \cite{Olney1996, Balaban2014, Kang2025}.
Beyond its clinical implications, this kinematic variability and asymmetry exacerbate distribution shifts for ML-driven exoskeleton controllers, which is particularly critical for controllers relying on sparse sensor observations such as the bilateral hip encoders used in this study.
To reproduce these conditions in able-bodied participants, we fitted a knee brace on the right knee to emulate stiff-knee gait (Fig. \ref{exo_hw_sw}a), and provided metronome cues that instructed participants to time their heel strikes to induce gait asymmetry (Fig. \ref{hemiplegic_gait_simulation}a).
The audio cues were set to align the impaired-side heel strike with the 75\% mark of the unimpaired side's gait cycle.
Stride durations were set to 1.4 and 1.1 seconds for the slow and nominal speeds used in the experimental protocol (Fig. \ref{Main_results}a).
Preliminary results from three participants confirmed that the emulated hemiplegic gait at the nominal speed deviated substantially from the able-bodied data, averaged across the 11 participants of the pre-training dataset, in both stance and swing durations.
The resulting swing time asymmetry ratio was 1.74 (Fig. \ref{hemiplegic_gait_simulation}b), closely matching the temporal characteristics of stroke survivors reported in previous studies \cite{Olney1996, Balaban2014}.
Lower-limb kinematics at the hip, knee, and ankle joints also showed clear deviations and asymmetry compared to able-bodied patterns.
To ensure that participants adapted properly to the emulated gait, they completed an adaptation period of at least five minutes across various speeds before the main experiment.

\subsection{Experimental Protocol}
We recruited nine able-bodied participants (7 male, 2 female) with a body mass of 68.3 $\pm$ 12.3 kg and a height of 173.1 $\pm$ 4.7 cm.
\rev{Three of these participants had also contributed to the 11-participant pre-training dataset, two to the pilot recordings used to select the update frequency (Fig.~\ref{effect_of_update_param}a, $n=4$), and one to those used to select the maximum replay number (Fig.~\ref{effect_of_update_param}b, $n=2$).
The pre-training data were recorded during unimpaired walking, whereas the experiments used emulated hemiplegic gait.}
The experimental protocol included six locomotor tasks with varying speeds and inclines (Fig. \ref{Main_results}a).
Participants first performed all six unseen tasks for 30 seconds each, so that the gait phase estimator was personalized to every task for the target user (Tasks 1--6).
The controller then underwent \rev{sustained single-task adaptation} for five minutes to induce a forgetting scenario, \rev{either LG slow or LG nominal} (Task 7\rev{, 72 or 91 consecutive updates, respectively}).
\rev{In daily ambulation settings, although walking occurs mostly in short bouts \cite{Orendurff2008, MoulaeeConradsson2025}, OA operates only while walking, so these bouts form one continuous stream of updates dominated by the most frequent task.}
Finally, the controller was tested on the remaining five tasks for 10 seconds each (Task 8), yielding five forgetting-testing pairs.
Although the task conditions were run as separate trials, the network weights and replay buffers were transferred from each trial to the next; both forgetting trials started from the state at the end of Task 6, and each testing trial started from the state at the end of the preceding forgetting trial.
Every trial began with 10 seconds of gait stabilization before the participant reached steady-state walking and adaptation began.
Each participant completed the same protocol for both OA without replay and \rev{gait feature-aware} replay, with the framework order randomized across participants.

\rev{To compare against the reservoir baselines, we conducted a replay comparison experiment with three able-bodied participants (all male, 72.3 $\pm$ 7.8 kg, and 171.7 $\pm$ 4.0 cm), one of whom also contributed to the pre-training dataset, the main experiment, and the pilot recordings for the update frequency.
The protocol was the same as in the main experiment, except that only the LG slow forgetting trial was used, extended to ten minutes (143 consecutive updates per limb), followed by its five testing trials.
Each participant completed this protocol with all four frameworks in Table~\ref{baselines} in randomized order.}

\subsection{\rev{Evaluation Metrics}}
\label{sec:metrics}
\rev{For each testing trial and limb, gait phase error was evaluated over the first 10~s of the trial, from the first detected heel strike to the last retained one, excluding the last two detected heel strikes in that window (Supplementary Material S1, Table~S1).
With $\hat{\phi}_k$ the estimated and $\phi_k$ the ground-truth gait phase (in \%, linearly interpolated between consecutive heel strikes) at time step $k$, the wrapped error is $e_k = \left((\phi_k - \hat{\phi}_k + 50) \bmod 100\right) - 50$, and
\begin{equation}
\text{RMSE}_{\phi} = \sqrt{\frac{1}{N}\sum_{k=1}^{N} e_k^2} \quad [\%].
\end{equation}
The torque command error compares the torque command sent to the actuator, $u_k$, with a reference torque $u^{\mathrm{ref}}_k$ obtained by using the ground-truth gait phase, task context, and the mid-level torque profile (Fig. \ref{exo_hw_sw}c):
\begin{equation}
\text{RMSE}_{\tau} = \sqrt{\frac{1}{N}\sum_{k=1}^{N} \left(u^{\mathrm{ref}}_k - u_k\right)^2} \quad [\text{N-m}].
\end{equation}
It quantifies the torque-profile error caused by gait phase estimation errors.
Both metrics were averaged over the two limbs, then over the two testing trials of each forgetting-testing pair (one after each forgetting trial), giving one value per participant and pair, and finally averaged over participants.
Percentage reductions are reported relative to OA without replay, $100 \times (\bar{E}_{\mathrm{no\ replay}} - \bar{E}_{\mathrm{replay}}) / \bar{E}_{\mathrm{no\ replay}}$, computed from the participant means.
Supplementary Material S1 details each step, verifies the reported values, and gives a worked example (Fig.~S1).}

\subsection{Statistical Analysis}
\rev{For the main experiment, a two-way repeated-measures analysis of variance (ANOVA) with adaptation method (without replay, proposed) and forgetting-testing pair (five levels) as within-subject factors was performed separately for torque command and gait phase RMSE.
Greenhouse--Geisser corrections were applied to effects with more than one degree of freedom.
Post-hoc paired t-tests compared the methods within each pair, with Bonferroni correction over the five pairs, and the pair-averaged values were compared with a paired t-test.
We report effect sizes as partial $\eta^2$ and Cohen's $d_z$, and 95\% confidence intervals of paired differences.
Because the replay comparison experiment included three participants, its results are reported descriptively without statistical tests.}

\section{Results}

\begin{figure}
    \centering
    \includegraphics[width=0.49\textwidth]{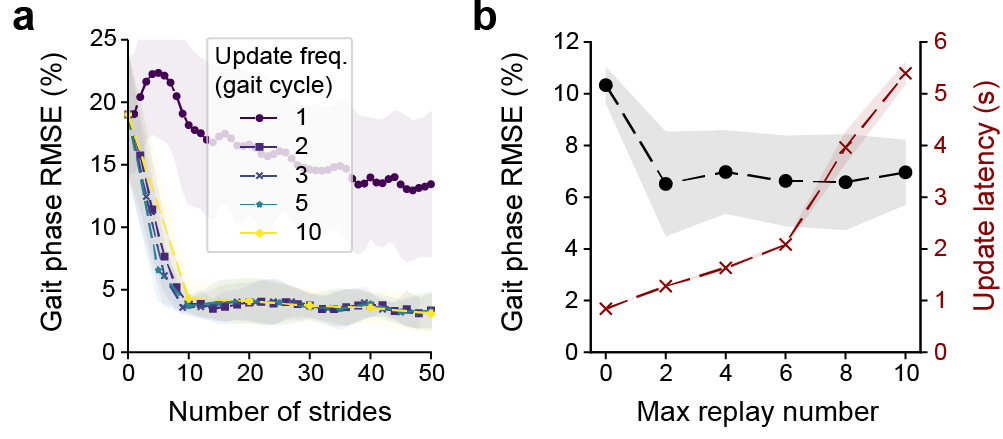}
    \caption{
        Effect of update parameters in offline data streaming simulation.
        (a) Effect of update frequency on gait phase error (mean $\pm$ SD, $n=4$).
        (b) Effect of maximum replay number on gait phase error and update latency (mean $\pm$ SD, $n=2$).
        }
    \label{effect_of_update_param}
\end{figure}

\subsection{Effect of Update Parameters on Online Adaptation}
For a fixed number of strides, we analyzed the effect of update frequency on estimator performance in an offline environment that mirrored the online control frequency and streamed pre-recorded emulated hemiplegic gait data.
Update frequencies greater than one gait cycle yielded similar accuracy, whereas updating every gait cycle slowed convergence considerably (Fig. \ref{effect_of_update_param}a).
This was primarily due to the 1-second TCN window, which limited the usable training data to the duration remaining after one second was subtracted from each gait cycle.
Consequently, although a one-cycle update is the most frequent, it produced the greatest convergence delay.
We therefore chose an update frequency of two gait cycles to ensure rapid convergence, triggering OA every third heel strike so that the system retained sufficient processing time by skipping one gait cycle after each pair.

\begin{figure*}[t]
    \centering
    \includegraphics[width=\textwidth]{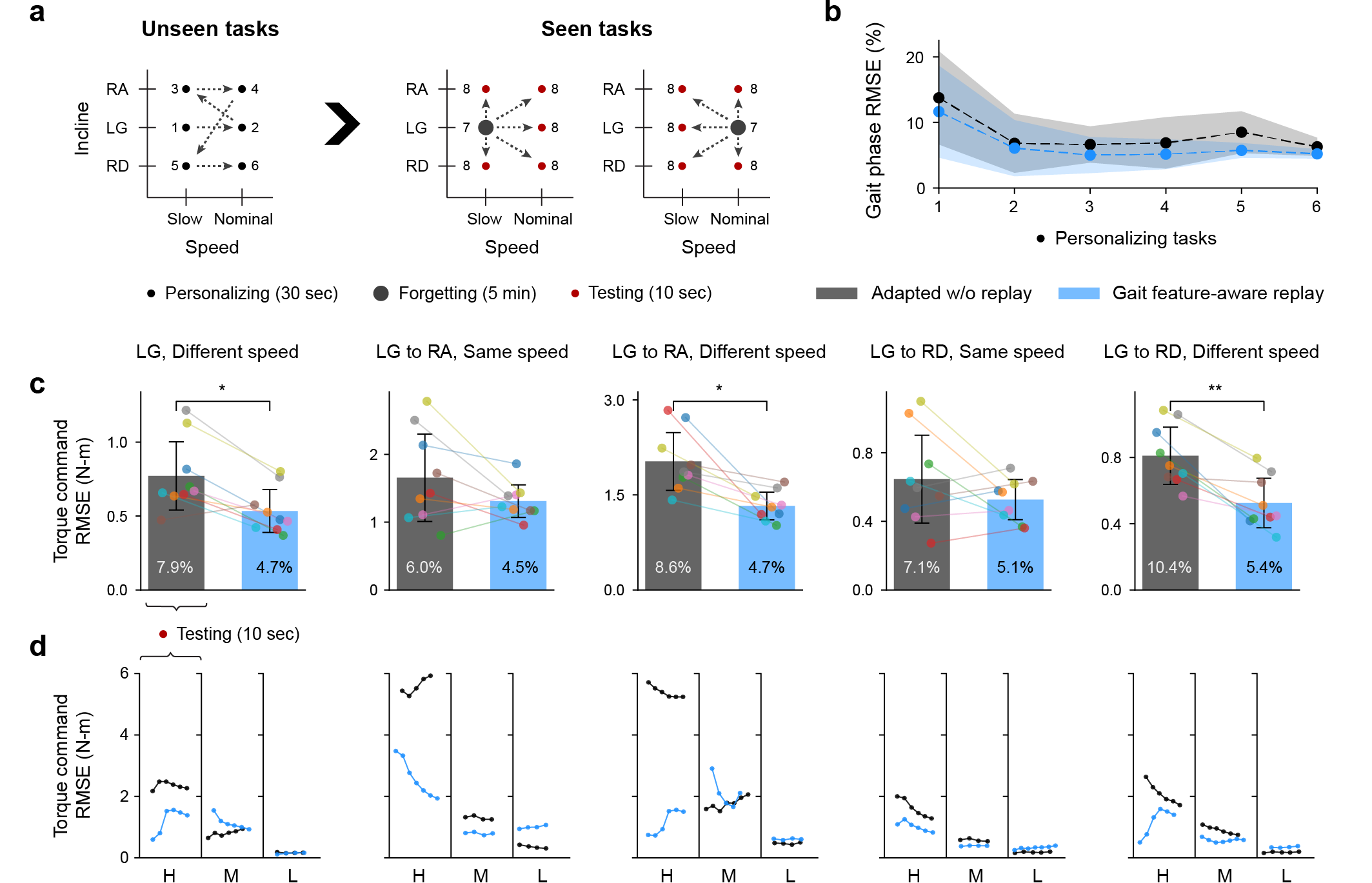}
    \caption{
        Experimental protocol and the results.
        (a) Experimental protocol across six locomotor tasks with the 1) personalization (Tasks 1--6), 2) forgetting (Task 7), and 3) testing (Task 8) phases.
        (b) Gait phase RMSE during the personalization phase across novel locomotor tasks (mean $\pm$ SD, $n=9$).
        (c) \rev{Torque command RMSE} across five forgetting-testing pairs during the testing phase.
Colored markers represent individual participants.
Asterisks denote \rev{significant post-hoc comparisons following the two-way repeated-measures ANOVA (Bonferroni-corrected paired t-tests;} *$p < 0.05$, **$p < 0.01$).
Inscribed numeric values on the bar graphs indicate the corresponding mean gait phase RMSE for each condition.
        (d) Representative forgetting scenarios for the five forgetting-testing pairs.
Subplots categorize the highest (H), median (M), and lowest (L) \rev{torque command RMSE} cases across all participants, tasks, and limbs.
Each marker corresponds to a single gait cycle \rev{within the evaluated span of} the 10-second testing window.
        }
    \label{Main_results}
\end{figure*}

We also optimized the maximum number of replays.
Because data loading and backpropagation time scale with the replay number, we selected a value that yielded update accuracy comparable to higher replay numbers while maintaining sufficiently low update latency, defined as the interval between the update trigger and the arrival of new network parameters at the main loop.
We evaluated estimator performance and update latency using the same protocol (Fig. \ref{Main_results}a), with data pre-recorded during pilot experiments.
A maximum of four replayed samples kept the update latency under two seconds while maintaining accuracy comparable to higher numbers, so we adopted that value (Fig. \ref{effect_of_update_param}b).

\subsection{Main Experiment}

Using the protocol described in Fig. \ref{Main_results}a, the main experiment showed that both OA without replay and \rev{gait feature-aware} replay began with a high gait phase error around 15\% during the first task of the personalization phase (Fig. \ref{Main_results}b).
After the initial personalizing task, errors for both cases stabilized under 10\%.
Throughout the personalization phase, gait phase RMSE across all novel tasks was 1.7 \rev{percentage points} lower with \rev{gait feature-aware} replay than with OA without replay, at \rev{$6.50 \pm 2.74\%$ and $8.23 \pm 3.03\%$}, respectively.

To analyze performance during the testing phase, we evaluated five different forgetting-testing task pairs: 1) LG, different speed, 2) LG to RA, same speed, 3) LG to RA, different speed, 4) LG to RD, same speed, and 5) LG to RD, different speed.
Each pair comprised two testing trials from both LG slow and LG nominal forgetting cases.
\rev{Torque command and gait phase RMSE were computed as defined in Section~\ref{sec:metrics}.}
For each of the five task pairs, values were averaged across the testing duration, both limbs, and the two testing trials (Fig.~\ref{Main_results}c).

\rev{Averaged over the five pairs, gait feature-aware replay reduced torque command RMSE by 28.6\% compared to OA without replay ($0.84 \pm 0.14$ vs. $1.18 \pm 0.27$~N-m; paired $t(8)=5.38$, $p<0.001$, $d_z=1.79$; Table~S2 for each participant).
In the two-way repeated-measures ANOVA (Greenhouse--Geisser-corrected degrees of freedom; full ANOVA tables in Table~S3), the main effects of method ($F(1,8)=28.9$, $p<0.001$, partial $\eta^2=0.78$) and pair ($F(2.21,17.65)=67.5$, $p<0.001$) were significant, whereas the method $\times$ pair interaction was not significant ($F(2.21,17.71)=2.98$, $p=0.073$).
For the five pairs, error reductions were 30.7\%, 20.8\%, 34.6\%, 18.5\%, and 35.2\%, respectively (Table~S4).
Significant improvements were observed in the (LG, different speed) and (LG to RA, different speed) pairs ($p=0.011$ and $p=0.022$, respectively), and the strongest effect was found in the (LG to RD, different speed) pair ($p=0.002$); the same-speed pairs were not significant (LG to RA: $p=0.61$; LG to RD: $p=1.00$).}
\rev{Gait feature-aware} replay exhibited a higher \rev{torque command} RMSE than OA without replay in only 1, 3, 0, 5, and 0 participants across the five task pairs.
\rev{Even in the task pairs with 1, 3, and 5 participants, the differences, 0.10, 0.27, and 0.09 N-m, respectively, were small relative to the overall error scale.}

\rev{For gait phase RMSE, gait feature-aware replay reduced the error by 39.2\% ($4.87 \pm 0.72\%$ vs. $8.00 \pm 1.48\%$; paired $t(8)=5.97$, $p<0.001$, $d_z=1.99$), with reductions of 40.3\%, 25.0\%, 45.4\%, 28.7\%, and 48.4\% for the five pairs (Table~S5 and Fig.~S2).
The main effects of method ($F(1,8)=35.6$, $p<0.001$, partial $\eta^2=0.82$) and pair ($F(2.36,18.87)=3.53$, $p=0.043$) were significant, without a significant method $\times$ pair interaction ($F(2.57,20.55)=1.97$, $p=0.16$).
Post-hoc tests were significant for the (LG, different speed) and (LG to RD, different speed) pairs ($p=0.036$ and $p=0.020$) and not for the (LG to RA, same speed), (LG to RA, different speed), and (LG to RD, same speed) pairs ($p=0.30$, $0.073$, and $0.41$).}
In the forgetting scenarios with the highest \rev{torque command} RMSE without replay among all participants, testing tasks, and limb sides, the error without replay consistently exceeded that with \rev{gait feature-aware} replay (H in Fig.~\ref{Main_results}d, which also shows the median (M) and lowest (L) cases).

\begin{figure}[t]
    \centering
    \includegraphics[width=0.49\textwidth]{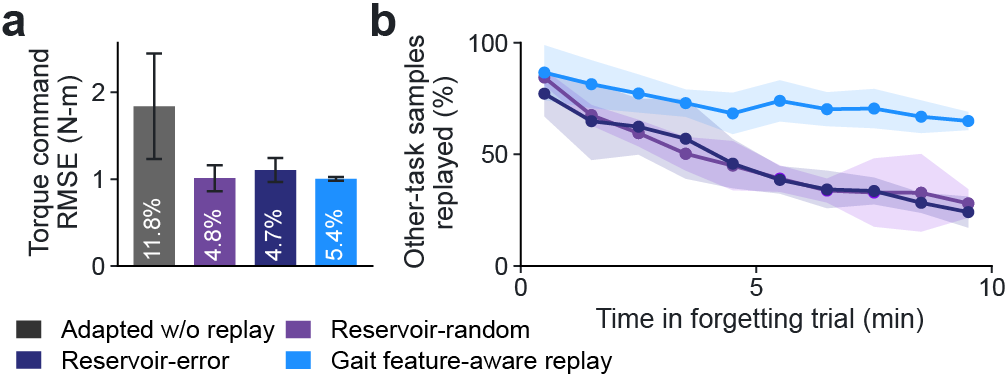}
    \caption{\rev{Comparison with the reservoir baselines ($n=3$).
(a) Torque command RMSE averaged over the five forgetting-testing pairs after the ten-minute LG slow forgetting trial (mean $\pm$ SD).
Inscribed numeric values indicate the corresponding mean gait phase RMSE for each framework.
(b) Share of the replayed samples that came from tasks other than the ongoing LG slow task during the forgetting trial, in one-minute bins with both limbs pooled (mean $\pm$ SD across participants).}}
    \label{baseline_results}
\end{figure}

\subsection{\rev{Comparison with Reservoir Baselines}}
\rev{Averaged over the forgetting-testing pairs, the three replay frameworks reduced torque command RMSE compared to OA without replay by a similar amount (Fig.~\ref{baseline_results}a): 45.1\% and 39.9\% for Reservoir-random and Reservoir-error, and 45.4\% for the gait feature-aware replay.
Gait phase RMSE was reduced by 59.0\%, 59.9\%, and 54.0\% for Reservoir-random, Reservoir-error, and the gait feature-aware replay, respectively.
Update latency during the forgetting trial was $1.7 \pm 0.1$~s for the gait feature-aware replay, $2.3 \pm 0.1$~s for both Reservoir-random and Reservoir-error, and $0.6 \pm 0.0$~s without replay.}

\rev{The storage policies differed in which tasks remained available for replay during the forgetting trial (Fig.~\ref{baseline_results}b).
In the first minute, 77--87\% of the replayed samples came from tasks other than the ongoing LG slow task in all three replay frameworks.
With the reservoir buffers, this share decreased steadily as LG slow samples accumulated, to $28 \pm 6\%$ (Reservoir-random) and $24 \pm 7\%$ (Reservoir-error) in the last minute, whereas with the gait feature-aware replay it decreased less, to $65 \pm 4\%$.
At the end of the forgetting trial, the replay grid still held $86.4 \pm 6.5\%$ of the other tasks' samples stored during the personalization phase, while Reservoir-random held $24.9 \pm 2.1\%$ and Reservoir-error held $24.4 \pm 1.3\%$.
Among the replayed samples of the non-ongoing tasks, the most-replayed task accounted for $31 \pm 6\%$ with Reservoir-random, $42 \pm 14\%$ with Reservoir-error, and $46 \pm 8\%$ with the gait feature-aware replay, where random retrieval distributed replay more evenly.

To examine conditions beyond the online experiments, we also replayed the recorded data of the replay comparison experiment through an open-loop offline copy of the controller (Supplementary Material S1, Section~S12).
When a fixed share of the replayed samples was drawn from the non-ongoing tasks, a 3\% share kept the torque command RMSE at its end-of-personalization level after the ten-minute forgetting trial (Fig.~S3).
When the forgetting trial was extended to three hours by repeating each participant's ten-minute LG slow recording, the torque command RMSE stayed at $0.86 \pm 0.16$~N-m with the gait feature-aware replay, whereas it rose to $1.28 \pm 0.20$ and $1.31 \pm 0.23$~N-m with Reservoir-random and Reservoir-error (Fig.~S4).
In the same simulation, building the replay grid on the two principal components only kept the torque command RMSE at $0.90 \pm 0.23$~N-m after three hours, whereas building it on gait cycle duration only raised it to $1.22 \pm 0.28$~N-m (Fig.~S5).
Storing up to 50 samples per limb in the gait feature space without a grid, with each new sample replacing its nearest stored sample, reached $0.85 \pm 0.06$~N-m after three hours, from $1.04 \pm 0.12$~N-m at the end of personalization.
}

\begin{figure}
    \centering
    \includegraphics[width=0.49\textwidth]{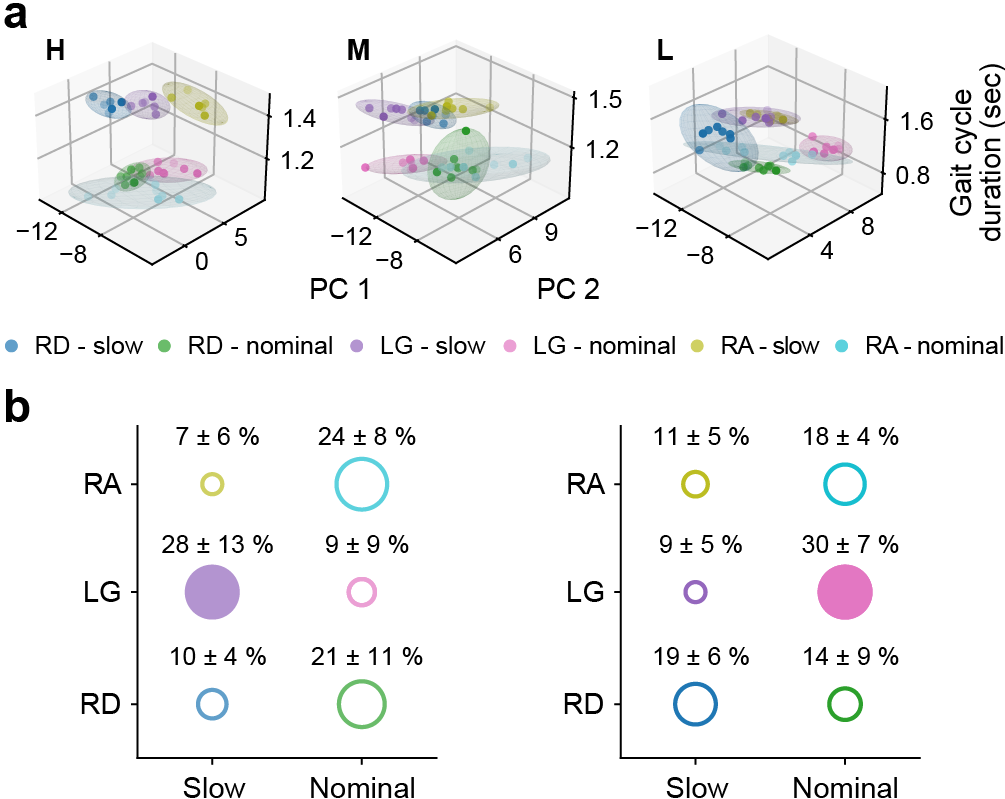}
    \caption{\rev{Gait feature space and replay distribution in the main experiment.}
        (a) \rev{Gait feature space} shown for \rev{the limbs} with the highest, median, and lowest silhouette scores\rev{ (0.22, 0.07, and $-0.05$)}, projected using the principal component extractor derived in Fig. \ref{gait_manifold}.
Each color indicates a distinct locomotor task, and ellipsoids represent the 90\% confidence intervals.
(b) Replay distributions during the forgetting phase for LG slow and LG nominal tasks.
Filled markers denote the ongoing tasks during the forgetting phase.
        }
    \label{replay_distribution}
\end{figure}

\subsection{Replay Distribution and Latency}

Post-hoc \rev{cluster analysis} revealed that the six locomotor tasks performed during the personalization phase formed \rev{partly} distinct representations within the \rev{gait feature space}.
\rev{Silhouette scores of individual limbs ranged from $-0.05$ to $0.22$ ($0.08 \pm 0.09$ across the 18 limbs, Fig.~\ref{replay_distribution}a), and participant-level scores, averaged over the two limbs, ranged from $-0.04$ to $0.18$.}
During each of the two five-minute forgetting trials, the ongoing task (LG slow or LG nominal) was replayed most frequently (Fig.~\ref{replay_distribution}b).
In the LG slow forgetting trial, RA nominal and RD nominal, which differed from the ongoing task in both speed and incline, were the next most frequently replayed (25\% and 21\% of the replayed samples). In the LG nominal forgetting trial, RD slow and RA nominal were replayed next (19\% and 18\%).
These distributions were computed retrospectively from the task labels recorded with each stored sample, which were not used to select the samples to replay during the online experiment.

\begin{figure}
    \centering
    \includegraphics[width=0.49\textwidth]{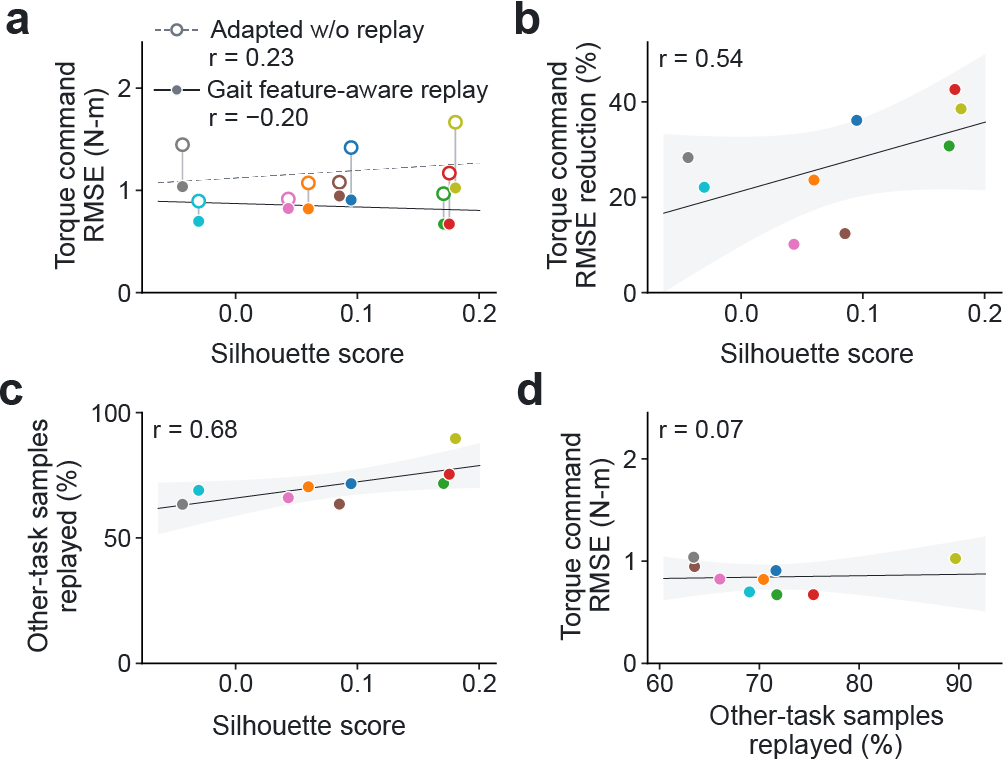}
    \caption{
        \rev{Effect of silhouette score on replay outcomes and on the replay of the non-ongoing tasks in the main experiment.
(a) Torque command RMSE of OA without replay (open markers, dashed line) and \rev{gait feature-aware} replay (filled markers, solid line) against the silhouette score; vertical lines connect the two values of each participant.
(b) Torque command RMSE reduction of \rev{gait feature-aware} replay relative to OA without replay, against the silhouette score.
(c) Share of the replayed samples that came from the non-ongoing tasks during the forgetting trials, against the silhouette score.
(d) Torque command RMSE of \rev{gait feature-aware} replay in (a) against the replayed share in (c).
Colors indicate participants as in Fig.~\ref{Main_results}. Lines show least-squares fits, shaded areas in (b)--(d) show their 95\% confidence bands, and $r$ is the Pearson correlation coefficient.}
        }
    \label{silhouette_analysis}
\end{figure}

\rev{To test whether participants with lower task separation benefited less from replay, we related each participant's silhouette score, averaged over the two limbs, to the outcomes of the main experiment (Fig.~\ref{silhouette_analysis}).
Across the nine participants, the torque command RMSE with replay remained similar regardless of the silhouette score ($r=-0.20$, Fig.~\ref{silhouette_analysis}a), as did the gait phase RMSE ($r=-0.08$).
However, without replay, both RMSEs tended to increase with the silhouette score ($r=0.23$ and $0.38$).
As a result, the reduction tended to increase with task separation, with $r=0.54$ for torque command RMSE (Fig.~\ref{silhouette_analysis}b) and $r=0.30$ for gait phase RMSE.
Participants with higher silhouette scores replayed a larger share of samples from the non-ongoing tasks during the forgetting trials ($r=0.68$, Fig.~\ref{silhouette_analysis}c).
Averaged over the two limbs and the two forgetting trials, this replayed share was $71.2 \pm 8.0\%$ across participants (range 63--90\%) and was correlated with the reduction in torque command RMSE ($r=0.68$) and gait phase RMSE ($r=0.51$), but not with the torque command and gait phase RMSE with replay ($r=0.07$, Fig.~\ref{silhouette_analysis}d, and $r=-0.27$).}

\begin{figure}
    \centering
    \includegraphics[width=0.49\textwidth]{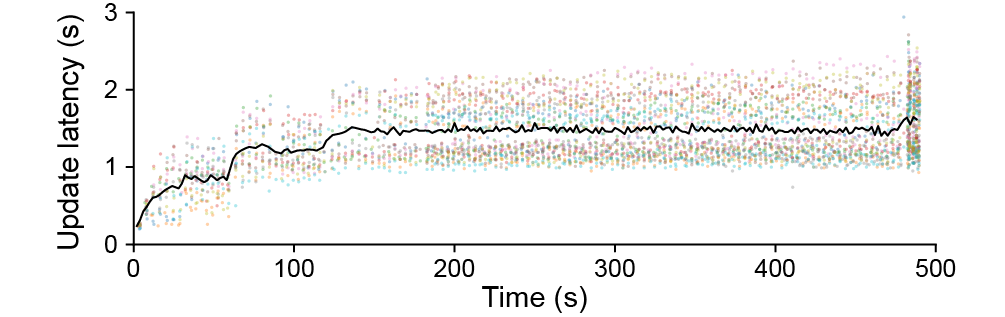}
    \caption{
        Update latency throughout the experiment.
The black line represents the average update latency across all participants, and each dot indicates the latency per individual update, with colors showing different participants.
        }
    \label{Replay_latency}
\end{figure}

After the exoskeleton had encountered approximately four tasks, the logged update latency plateaued at roughly 1.5 seconds, which corresponds to the maximum replay number of four (Fig.~\ref{Replay_latency}).
Throughout the plateau period, the update latency remained within a band of 1.5 $\pm$ 0.3 seconds, with peak latency never exceeding three seconds.

\section{Discussion}
In this work, we proposed a \rev{gait feature-aware} replay-based OA framework for rapid adaptation to \rev{emulated hemiplegic gait patterns}.
Compared to \rev{OA} without experience replay, \rev{gait feature-aware} replay reduced \rev{gait phase RMSE by 39.2\% and torque command RMSE by 28.6\% in able-bodied participants emulating hemiplegic gait}.
OA without replay was susceptible to catastrophic forgetting of previously experienced tasks due to overfitting on the current task, whereas \rev{gait feature-aware} replay revisited previously seen tasks without requiring locomotor context labels, an advantage for real-world deployments where such labels are difficult to obtain.
This forgetting was strongest in test pairs involving different speeds, more so than incline transitions alone, likely due to larger kinematic differences between speeds than between inclines.
\rev{However, even for the same-speed incline pairs, six and four of the nine participants exhibited higher torque command RMSE without replay for LG to RA and LG to RD, respectively.}
The average update latency was 1.5 seconds, so the network \rev{could respond quickly to changes in the user's gait}.
\rev{In the replay comparison experiment, reservoir replay with the same replay budget reduced errors to a similar extent, so the error reduction over OA without replay is attributable mainly to replay itself, while the replay grid kept more of the non-ongoing tasks available for replay during prolonged single-task walking.}

\rev{The reservoir baselines test whether the benefit comes from replay in general or from how the buffer is organized and sampled.
Because reservoir sampling preserved a uniform sample of the stream, only 24.9\% and 24.4\% of the other tasks' samples stored during the personalization phase remained in the Reservoir-random and Reservoir-error buffers after the ten-minute forgetting trial.
The gait feature-aware replay instead kept 86\% of these samples, and the non-ongoing tasks remained the majority of the replayed samples (Fig.~\ref{baseline_results}b).
The reservoir baselines also had a longer update latency (2.3 vs. 1.7~s for the gait feature-aware replay) during the forgetting trial.
This was because the reservoir buffers were full at the 50-sample budget, whereas the replay grid held 22--45 occupied cells per limb during this trial, so each update evaluated more stored samples with the reservoir buffers.
This gap could widen in longer deployments, in which a reservoir would need a larger budget to retain rarely performed tasks, whereas the replay grid grows only with the region of the gait feature space that the user's gait covers.
This is consistent with the offline simulation of a 180-minute forgetting trial in Fig.~S4, where the reservoir baselines gradually forgot as their buffers filled with the ongoing task, whereas gait feature-aware replay did not.
In the offline ablation (Fig.~S5), the principal components provided most of this retention, whereas gait cycle duration alone formed only about five cells per limb and forgot similarly to the reservoir baselines.
A 50-sample buffer that replaced the nearest stored sample without a grid also prevented forgetting, so the retention came from storing samples by their location in the gait feature space rather than from the discretization, which bounds the buffer without a fixed budget or comparisons with stored samples.}

\rev{Despite this difference, the gait feature-aware replay did not reduce errors beyond the reservoir baselines (Fig.~\ref{baseline_results}a), likely for two reasons.
First, all six tasks involved steady-state treadmill walking with overlapping gait features (Fig.~\ref{replay_distribution}a). A buffer dominated by LG slow samples yet retaining a small share of the other tasks may therefore have sufficed to prevent forgetting.
In offline simulations (Fig.~S3), a 3\% share of non-ongoing tasks proved sufficient to prevent forgetting.
Second, gait phase is defined identically for every task as the normalized progression between consecutive heel strikes, so its mapping from hip kinematics is also shared across tasks.
Therefore, organizing the buffer by gait features should matter more when the mapping from sensor inputs to the estimated user state largely differs across tasks, such as biological joint torque \cite{Molinaro2024, Molinaro2024SciRob}.
Likewise, clear feature separation becomes more critical when the task set exhibits high variability, including non-cyclic activities \cite{Molinaro2024}.
Under these conditions, a uniform buffer would progressively lose the less frequent tasks (Fig.~\ref{baseline_results}b) while the gait feature-aware replay keeps them as long as their features occupy separate regions.
Error-prioritized retrieval also yielded no measurable difference between the two reservoir baselines, even though it is widely used in continual learning \cite{Aljundi2019MIR, schaul2016prioritized}. We expect this to matter more in continual personalization scenarios with higher task variance, such as sets spanning both cyclic and non-cyclic tasks.}

Tasks were heterogeneously replayed during the forgetting phase (Fig.~\ref{replay_distribution}b).
Although ongoing task data was integrated and replayed alongside the target replay tasks, the ongoing tasks were replayed most frequently, likely reflecting a balancing mechanism that manages representation conflicts between task sets.
Tasks differing from the ongoing task in both speed and incline were mostly replayed next, consistent with their largest improvement over OA without replay.
\rev{Near-zero silhouette scores in Fig.~\ref{replay_distribution} and Fig.~\ref{silhouette_analysis} show that the gait feature space did not entirely separate tasks into distinct clusters.
Although the silhouette score was moderately correlated with the replayed share of the other tasks and with the error reduction, it showed no clear relation to the torque command and gait phase RMSE with replay across the nine participants (Fig.~\ref{silhouette_analysis}).
This may occur because tasks that share a cyclic walking pattern, such as walking at adjacent speeds and inclines, can overlap in certain grid cells while each task still preserves distinct, rarely shared cells of its own.
Their samples in a shared cell likely map sensor inputs to gait phase similarly, so overwriting the cell with the ongoing task loses little.
Accordingly, the RMSE with replay was not lower in participants who replayed more of the other tasks' samples (Fig.~\ref{silhouette_analysis}d).
Therefore, in this study's task set, the level of separation among adjacent tasks did not largely affect the error with replay, although the separation was associated with the replayed share (Fig.~\ref{silhouette_analysis}c).
With nine participants and participant-level silhouette scores spanning only $-0.04$ to $0.18$, these correlations are exploratory and require confirmation in larger cohorts.
However, when the mapping from sensor inputs to the estimated user state differs across tasks, an overwritten cell would lose task-specific information, and representations with better task separation, such as learned embeddings \cite{Pellegrini2020, Hayes2020, Yuan2024}, may then be needed.
}

Freezing the TCN blocks proved effective for preserving high-level, general gait features during OA.
This is confirmed in Fig. \ref{Main_results}b, where tasks following the initial task maintained lower gait phase errors, indicating that user-specific features in the final linear layer were updated.
Isolating the roles of individual layers in representing generic versus subject-specific gait remains difficult, as these representations are often deeply coupled, and the final linear layer is prone to overfitting on continuously streamed data, particularly during prolonged exposure to a single task.
Integrating experience replay with this architectural strategy is therefore essential for maintaining controller stability across a diverse task set.

Although this study introduces a \rev{task label-free replay buffer} for exoskeleton control, several limitations remain.
First, retrospective gait phase labeling relied on force plate data \rev{from an instrumented treadmill}.
In real-world applications, this labeling should be offloaded to portable sensing modalities such as pressure-sensing insoles or inertial measurement units.
\rev{Second, the controller still used task context from the experimental protocol to select the torque splines, so the complete controller is not entirely task-agnostic. Deploying it outside the laboratory would additionally require a task recognizer or task-independent assistance profiles.}
Third, this framework adapts a gait phase estimator, which primarily updates the high-level user state \rev{and therefore personalizes assistance timing}, while relying on mid-level control profiles derived from prior able-bodied studies \cite{Franks2022}.
\rev{The magnitude and shape of assistance were not personalized, and the torque command RMSE reflects deviations from these predefined profiles rather than measured actuator torque or user benefit.}
These profiles should also be personalized, particularly for clinical populations with high inter-subject variability.
This could be addressed by adopting other user state estimators such as biological joint moment estimators \cite{Molinaro2024, Molinaro2024SciRob}, which depend less on fixed mid-level parameters. However, retrospective labeling of user states remains a persistent challenge for supervised adaptation.
A promising alternative is reinforcement learning (RL)-based adaptation \cite{Zhong2025, Zhang2026}, which could tune mid-level control or learn end-to-end strategies, though rapid RL adaptation across diverse, unstructured tasks remains an open challenge in the exoskeleton field.

Fourth, this study used a fixed learning rate, which may not suit diverse gait transition scenarios that demand a balance between rapid adaptation to new tasks and long-term stability for previously experienced tasks.
Future work will explore scheduling mechanisms for both learning rates and replay counts by tracking training loss or other stability-plasticity metrics from the continual learning domain \cite{Johnson2025}, and will implement logic to suspend adaptation when further updates are unnecessary.
Finally, the target users in this study \rev{were able-bodied participants who} walked with an emulated hemiplegic gait \rev{on a treadmill under controlled transitions}, which \rev{reproduced the temporal asymmetry and stiff-knee kinematics of hemiplegic gait but} may not fully capture \rev{the neuromuscular impairment and variability of} real-world stroke survivors.
Future work will recruit patients to investigate biomechanical challenges such as increased intra-subject gait variability and its impact on OA performance\rev{, and will evaluate the framework during overground, self-paced community ambulation}.

\section{Conclusion}
\rev{We developed an experience replay framework for online adaptation of an exoskeleton gait phase estimator that organizes its replay buffer in a PCA-based gait feature space and therefore stores and selects replay data without locomotor task labels.
In able-bodied participants emulating hemiplegic gait under controlled treadmill transitions, it reduced gait phase RMSE by 39.2\% and torque command RMSE by 28.6\% compared to online adaptation without replay.
These results support task label-free replay for stabilizing gait phase estimator personalization across locomotor transitions.
We expect that this framework will allow continual exoskeleton personalization across daily locomotor tasks once validated in clinical populations and combined with portable heel strike labeling and task-independent assistance profiles.}

\section*{Acknowledgments}
The authors would like to thank \anon{Weijie Lim, Cole Drake, Atri Dey, and David Newsom}{xxxx} who helped with the data collection, everyone who participated in the experiments, all the students in the hip exoskeleton team, \anon{Maria Tagliaferri, Jimin An, Rajiv Joshi, and Yunliang Zhao}{xxxx}, who were involved in developing the hardware.

\bibliographystyle{IEEEtran}
\bibliography{references}

@article{sawicki2020,
  title={The exoskeleton expansion: improving walking and running economy},
  author={Sawicki, Gregory S and Beck, Owen N and Kang, Inseung and Young, Aaron J},
  journal={Journal of neuroengineering and rehabilitation},
  volume={17},
  number={1},
  pages={25},
  year={2020},
  publisher={Springer}
}

@article{siviy2023,
  title={Opportunities and challenges in the development of exoskeletons for locomotor assistance},
  author={Siviy, Christopher and others},
  journal={Nature biomedical engineering},
  volume={7},
  number={4},
  pages={456--472},
  year={2023},
  publisher={Nature Publishing Group UK London}
}

@article{Gao2025,
  title = {Wearable technologies for assisted mobility in the real world},
  volume = {16},
  ISSN = {2041-1723},
  DOI = {10.1038/s41467-025-67126-4},
  number = {1},
  journal = {Nature Communications},
  publisher = {Springer Science and Business Media LLC},
  author = {Gao,  Shuo and others},
  year = {2025},
  month = dec 
}

@article{Awad2017,
  title = {A soft robotic exosuit improves walking in patients after stroke},
  volume = {9},
  ISSN = {1946-6242},
  DOI = {10.1126/scitranslmed.aai9084},
  number = {400},
  journal = {Science Translational Medicine},
  publisher = {American Association for the Advancement of Science (AAAS)},
  author = {Awad,  Louis N. and others},
  year = {2017},
  month = jul 
}

@article{Pruyn2026,
  title = {Portable hip exoskeleton improves walking economy for stroke survivors},
  volume = {17},
  ISSN = {2041-1723},
  DOI = {10.1038/s41467-026-69580-0},
  number = {1},
  journal = {Nature Communications},
  publisher = {Springer Science and Business Media LLC},
  author = {Pruyn,  Kai and Murray,  Rosemarie and Gabert,  Lukas and Foreman,  K. Bo and Lenzi,  Tommaso},
  year = {2026},
  month = feb 
}

@article{Archangeli2026,
  title = {Exoskeleton frontal and sagittal plane hip torque improves propulsion and transient stability during walking in individuals with hemiparesis},
  ISSN = {1743-0003},
  DOI = {10.1186/s12984-026-01937-4},
  journal = {Journal of NeuroEngineering and Rehabilitation},
  publisher = {Springer Science and Business Media LLC},
  author = {Archangeli,  Dante and others},
  year = {2026},
  month = May 
}

@article{Lerner2017SciTranslMed,
    author = {Zachary F. Lerner  and Diane L. Damiano  and Thomas C. Bulea },
    title = {A lower-extremity exoskeleton improves knee extension in children with crouch gait from cerebral palsy},
    journal = {Science Translational Medicine},
    volume = {9},
    number = {404},
    pages = {eaam9145},
    year = {2017},
    doi = {10.1126/scitranslmed.aam9145},
    eprint = {https://www.science.org/doi/pdf/10.1126/scitranslmed.aam9145},
}

@inproceedings{Divekar2025,
  title = {A Task-Agnostic Knee Exoskeleton for Reducing Osteoarthritis Pain Across Activities of Daily Life: A Pilot Study},
  DOI = {10.1109/icorr66766.2025.11063102},
  booktitle = {2025 International Conference On Rehabilitation Robotics (ICORR)},
  publisher = {IEEE},
  author = {Divekar,  Nikhil V. and Hinojosa,  Ernesto Hernandez and Zhang,  Jiefu and Gregg,  Robert D.},
  year = {2025},
  month = May,
  pages = {1437–1443}
}

@inproceedings{Zhang2025,
  title = {A Task-Agnostic Hip Exoskeleton for Osteoarthritis Pain Relief: Energetic Control Across Activities of Daily Life},
  DOI = {10.1109/icorr66766.2025.11063157},
  booktitle = {2025 International Conference On Rehabilitation Robotics (ICORR)},
  publisher = {IEEE},
  author = {Zhang,  Jiefu and Divekar,  Nikhil V. and Hinojosa,  Ernesto Hernandez and Gregg,  Robert D.},
  year = {2025},
  month = May,
  pages = {1299–1306}
}

@article{Ingraham2022,
  title = {The role of user preference in the customized control of robotic exoskeletons},
  volume = {7},
  ISSN = {2470-9476},
  DOI = {10.1126/scirobotics.abj3487},
  number = {64},
  journal = {Science Robotics},
  publisher = {American Association for the Advancement of Science (AAAS)},
  author = {Ingraham,  K. A. and Remy,  C. D. and Rouse,  E. J.},
  year = {2022},
  month = Mar 
}

@article{Lee2023,
  title = {User preference optimization for control of ankle exoskeletons using sample efficient active learning},
  volume = {8},
  ISSN = {2470-9476},
  DOI = {10.1126/scirobotics.adg3705},
  number = {83},
  journal = {Science Robotics},
  publisher = {American Association for the Advancement of Science (AAAS)},
  author = {Lee,  Ung Hee and others},
  year = {2023},
  month = Oct 
}

@article{molinaro2024,
  title={Task-agnostic exoskeleton control via biological joint moment estimation},
  author={Molinaro, Dean D and others},
  journal={Nature},
  volume={635},
  number={8038},
  pages={337--344},
  year={2024},
  publisher={Nature Publishing Group UK London}
}

@article{Winter1984,
  title = {Kinematic and kinetic patterns in human gait: Variability and compensating effects},
  volume = {3},
  ISSN = {0167-9457},
  DOI = {10.1016/0167-9457(84)90005-8},
  number = {1–2},
  journal = {Human Movement Science},
  publisher = {Elsevier BV},
  author = {Winter,  David A.},
  year = {1984},
  month = mar,
  pages = {51–76}
}

@article{Balaban2014,
  title = {Gait Disturbances in Patients With Stroke},
  volume = {6},
  ISSN = {1934-1563},
  DOI = {10.1016/j.pmrj.2013.12.017},
  number = {7},
  journal = {PM\&R},
  publisher = {Wiley},
  author = {Balaban,  Birol and Tok,  Fatih},
  year = {2014},
  month = jan,
  pages = {635–642}
}

@article{Olney1996,
  title = {Hemiparetic gait following stroke. Part I: Characteristics},
  volume = {4},
  ISSN = {0966-6362},
  DOI = {10.1016/0966-6362(96)01063-6},
  number = {2},
  journal = {Gait \& Posture},
  publisher = {Elsevier BV},
  author = {Olney,  Sandra J. and Richards,  Carol},
  year = {1996},
  month = Apr,
  pages = {136–148}
}

@article{Kang2025,
  title = {Online Adaptation Framework Enables Personalization of Exoskeleton Assistance During Locomotion in Patients Affected by Stroke},
  volume = {41},
  ISSN = {1941-0468},
  DOI = {10.1109/tro.2025.3595701},
  journal = {IEEE Transactions on Robotics},
  publisher = {Institute of Electrical and Electronics Engineers (IEEE)},
  author = {Kang,  Inseung and Molinaro,  Dean D. and Park,  Dongho and Lee,  Dawit and Kunapuli,  Pratik and Herrin,  Kinsey R. and Young,  Aaron J.},
  year = {2025},
  pages = {4941–4959}
}

@article{Gunnell2025,
  title = {Powered knee exoskeleton improves sit-to-stand transitions in stroke patients using electromyographic control},
  volume = {4},
  ISSN = {2731-3395},
  DOI = {10.1038/s44172-025-00440-3},
  number = {1},
  journal = {Communications Engineering},
  publisher = {Springer Science and Business Media LLC},
  author = {Gunnell,  Andrew J. and others},
  year = {2025},
  month = jun 
}

@article{Kim2024,
  title = {Soft robotic apparel to avert freezing of gait in Parkinson’s disease},
  volume = {30},
  ISSN = {1546-170X},
  DOI = {10.1038/s41591-023-02731-8},
  number = {1},
  journal = {Nature Medicine},
  publisher = {Springer Science and Business Media LLC},
  author = {Kim,  Jinsoo and others},
  year = {2024},
  month = jan,
  pages = {177–185}
}

@article{Slade2022personalizing,
  title = {Personalizing exoskeleton assistance while walking in the real world},
  volume = {610},
  ISSN = {1476-4687},
  DOI = {10.1038/s41586-022-05191-1},
  number = {7931},
  journal = {Nature},
  publisher = {Springer Science and Business Media LLC},
  author = {Slade,  Patrick and Kochenderfer,  Mykel J. and Delp,  Scott L. and Collins,  Steven H.},
  year = {2022},
  month = oct,
  pages = {277–282}
}

@INPROCEEDINGS{Johnson2023,
  author={Johnson, C. and Cho, J. and Maldonado-Contreras, J. and Chaluvadi, S. and Young, A. J.},
  booktitle={2023 International Symposium on Medical Robotics (ISMR)}, 
  title={Adaptive Lower-Limb Prosthetic Control: Towards Personalized Intent Recognition \& Context Estimation}, 
  year={2023},
  volume={},
  number={},
  pages={1-7},
  doi={10.1109/ISMR57123.2023.10130251}}

@ARTICLE{Maldonado2025,
  author={Maldonado-Contreras, Jairo Y. and others},
  journal={IEEE Transactions on Medical Robotics and Bionics}, 
  title={Real-Time Adaptation of Deep Learning Walking Speed Estimators Enables Biomimetic Assistance Modulation in an Open-Source Bionic Leg}, 
  year={2025},
  volume={7},
  number={2},
  pages={711-722},
  doi={10.1109/TMRB.2025.3550642}}

@ARTICLE{Johnson2025,
  author={Johnson, Cole B. and Maldonado-Contreras, Jairo and Herrin, Kinsey R. and Young, Aaron J.},
  journal={IEEE Transactions on Medical Robotics and Bionics}, 
  title={Real-Time Balancing of Stability and Plasticity in Continual Learning Enables Adaptive Speed Estimation for Lower-Limb Prostheses}, 
  year={2025},
  volume={7},
  number={4},
  pages={1634-1645},
  doi={10.1109/TMRB.2025.3597014}}

@article{Young2017,
  title = {Influence of Power Delivery Timing on the Energetics and Biomechanics of Humans Wearing a Hip Exoskeleton},
  volume = {5},
  ISSN = {2296-4185},
  DOI = {10.3389/fbioe.2017.00004},
  journal = {Frontiers in Bioengineering and Biotechnology},
  publisher = {Frontiers Media SA},
  author = {Young,  Aaron J. and Foss,  Jessica and Gannon,  Hannah and Ferris,  Daniel P.},
  year = {2017},
  month = Mar 
}

@article{Luo2024,
  title = {Experiment-free exoskeleton assistance via learning in simulation},
  volume = {630},
  ISSN = {1476-4687},
  DOI = {10.1038/s41586-024-07382-4},
  number = {8016},
  journal = {Nature},
  publisher = {Springer Science and Business Media LLC},
  author = {Luo,  Shuzhen and others},
  year = {2024},
  month = jun,
  pages = {353–359}
}

@misc{Park2026,
  doi = {10.48550/ARXIV.2603.04166},
  author = {Park,  Ilseung and Song,  Changseob and Kang,  Inseung},
  title = {Learning Hip Exoskeleton Control Policy via Predictive Neuromusculoskeletal Simulation},
  publisher = {arXiv},
  year = {2026},
  copyright = {Creative Commons Attribution 4.0 International}
}

@ARTICLE{Kang2021,
  author={Kang, Inseung and others},
  journal={IEEE Robotics and Automation Letters}, 
  title={Real-Time Gait Phase Estimation for Robotic Hip Exoskeleton Control During Multimodal Locomotion}, 
  year={2021},
  volume={6},
  number={2},
  pages={3491-3497},
  doi={10.1109/LRA.2021.3062562}}

@ARTICLE{Kang2019,
  author={Kang, Inseung and Kunapuli, Pratik and Young, Aaron J.},
  journal={IEEE Transactions on Medical Robotics and Bionics}, 
  title={Real-Time Neural Network-Based Gait Phase Estimation Using a Robotic Hip Exoskeleton}, 
  year={2020},
  volume={2},
  number={1},
  pages={28-37},
  doi={10.1109/TMRB.2019.2961749}}

@article{Molinaro2024SciRob,
  title = {Estimating human joint moments unifies exoskeleton control,  reducing user effort},
  volume = {9},
  ISSN = {2470-9476},
  DOI = {10.1126/scirobotics.adi8852},
  number = {88},
  journal = {Science Robotics},
  publisher = {American Association for the Advancement of Science (AAAS)},
  author = {Molinaro,  Dean D. and Kang,  Inseung and Young,  Aaron J.},
  year = {2024},
  month = mar 
}

@article{Lee2024,
  title = {AI-driven universal lower-limb exoskeleton system for community ambulation},
  volume = {10},
  ISSN = {2375-2548},
  DOI = {10.1126/sciadv.adq0288},
  number = {51},
  journal = {Science Advances},
  publisher = {American Association for the Advancement of Science (AAAS)},
  author = {Lee,  Dawit and Lee,  Sanghyub and Young,  Aaron J.},
  year = {2024},
  month = dec 
}

@ARTICLE{Franks2022,
  author={Franks, Patrick W. and others},
  journal={IEEE Transactions on Neural Systems and Rehabilitation Engineering}, 
  title={The Effects of Incline Level on Optimized Lower-Limb Exoskeleton Assistance: A Case Series}, 
  year={2022},
  volume={30},
  number={},
  pages={2494-2505},
  doi={10.1109/TNSRE.2022.3196665}}

@article{Kang2022,
  title = {Subject-Independent Continuous Locomotion Mode Classification for Robotic Hip Exoskeleton Applications},
  volume = {69},
  ISSN = {1558-2531},
  DOI = {10.1109/tbme.2022.3165547},
  number = {10},
  journal = {IEEE Transactions on Biomedical Engineering},
  publisher = {Institute of Electrical and Electronics Engineers (IEEE)},
  author = {Kang,  Inseung and Molinaro,  Dean D. and Choi,  Gayeon and Camargo,  Jonathan and Young,  Aaron J.},
  year = {2022},
  month = oct,
  pages = {3234–3242}
}

@article{Wang2024,
  title = {A Comprehensive Survey of Continual Learning: Theory,  Method and Application},
  volume = {46},
  ISSN = {1939-3539},
  DOI = {10.1109/tpami.2024.3367329},
  number = {8},
  journal = {IEEE Transactions on Pattern Analysis and Machine Intelligence},
  publisher = {Institute of Electrical and Electronics Engineers (IEEE)},
  author = {Wang,  Liyuan and Zhang,  Xingxing and Su,  Hang and Zhu,  Jun},
  year = {2024},
  month = aug,
  pages = {5362–5383}
}

@article{Delange2021,
  title = {A continual learning survey: Defying forgetting in classification tasks},
  ISSN = {1939-3539},
  DOI = {10.1109/tpami.2021.3057446},
  journal = {IEEE Transactions on Pattern Analysis and Machine Intelligence},
  publisher = {Institute of Electrical and Electronics Engineers (IEEE)},
  author = {Delange,  Matthias and others},
  year = {2021},
  pages = {1–1}
}

@misc{Rolnick2018,
  doi = {10.48550/ARXIV.1811.11682},
  author = {Rolnick,  David and Ahuja,  Arun and Schwarz,  Jonathan and Lillicrap,  Timothy P. and Wayne,  Greg},
  title = {Experience Replay for Continual Learning},
  publisher = {arXiv},
  year = {2018},
  copyright = {arXiv.org perpetual,  non-exclusive license}
}

@ARTICLE{Mermillod2013,
    AUTHOR={Mermillod, Martial  and Bugaiska, Aurélia  and Bonin, Patrick},
    TITLE={The stability-plasticity dilemma: investigating the continuum from catastrophic forgetting to age-limited learning effects},
    JOURNAL={Frontiers in Psychology},
    VOLUME={Volume 4},
    YEAR={2013},
    DOI={10.3389/fpsyg.2013.00504},
    ISSN={1664-1078},
}

@misc{nvidia_tensorrt,
  author = {{NVIDIA}},
  title = {{NVIDIA TensorRT}},
  year = {2026},
  url = {https://developer.nvidia.com/tensorrt},
  note = {Accessed: April 27, 2026}
}

@incollection{PyTorch,
  title = {PyTorch: An Imperative Style, High-Performance Deep Learning Library},
  author = {Paszke, Adam and others},
  booktitle = {Advances in Neural Information Processing Systems 32},
  pages = {8024--8035},
  year = {2019},
}

@inproceedings{Yosinski2014,
 author = {Yosinski, Jason and Clune, Jeff and Bengio, Yoshua and Lipson, Hod},
 booktitle = {Advances in Neural Information Processing Systems},
 pages = {},
 title = {How transferable are features in deep neural networks?},
 volume = {27},
 year = {2014}
}

@article{Zhang2026,
  title = {Towards adaptive optimal personalization control of robotic hip exoskeleton assistance for individuals with paretic stroke},
  ISSN = {1558-2531},
  DOI = {10.1109/tbme.2026.3685273},
  journal = {IEEE Transactions on Biomedical Engineering},
  publisher = {Institute of Electrical and Electronics Engineers (IEEE)},
  author = {Zhang,  Qiang and Chen,  Yun and Lewek,  Michael and Si,  Jennie and Huang,  He},
  year = {2026},
  pages = {1–12}
}

@inproceedings{Zhong2025,
author = {Zhong, Junmin and others},
title = {Reinforcement learning control of a physical robot device for assisted human walking without a simulator},
year = {2025},
publisher = {JMLR.org},
booktitle = {Proceedings of the 42nd International Conference on Machine Learning},
articleno = {3164},
numpages = {27},
location = {Vancouver, Canada},
series = {ICML'25}
}

@inproceedings{Isele2018,
  author    = {Isele, David and Cosgun, Akansel},
  title     = {Selective Experience Replay for Lifelong Learning},
  booktitle = {Proceedings of the AAAI Conference on Artificial Intelligence},
  volume    = {32},
  number    = {1},
  pages     = {3302--3309},
  year      = {2018}
}

@inproceedings{Aljundi2019GSS,
  author    = {Aljundi, Rahaf and Lin, Min and Goujaud, Baptiste and Bengio, Yoshua},
  title     = {Gradient Based Sample Selection for Online Continual Learning},
  booktitle = {Advances in Neural Information Processing Systems},
  volume    = {32},
  year      = {2019}
}

@inproceedings{Aljundi2019MIR,
  author    = {Aljundi, Rahaf and Belilovsky, Eugene and Tuytelaars, Tinne and Charlin, Laurent and Caccia, Massimo and Lin, Min and Page-Caccia, Lucas},
  title     = {Online Continual Learning with Maximal Interfered Retrieval},
  booktitle = {Advances in Neural Information Processing Systems},
  volume    = {32},
  year      = {2019}
}

@inproceedings{Aljundi2019TaskFree,
  author    = {Aljundi, Rahaf and Kelchtermans, Klaas and Tuytelaars, Tinne},
  title     = {Task-Free Continual Learning},
  booktitle = {Proceedings of the IEEE/CVF Conference on Computer Vision and Pattern Recognition (CVPR)},
  pages     = {11254--11263},
  year      = {2019}
}

@misc{Chaudhry2019,
  author        = {Chaudhry, Arslan and Rohrbach, Marcus and Elhoseiny, Mohamed and Ajanthan, Thalaiyasingam and Dokania, Puneet K. and Torr, Philip H. S. and Ranzato, Marc'Aurelio},
  title         = {On Tiny Episodic Memories in Continual Learning},
  howpublished  = {arXiv preprint arXiv:1902.10486},
  year          = {2019}
}

@article{Vitter1985,
  author  = {Vitter, Jeffrey S.},
  title   = {Random Sampling with a Reservoir},
  journal = {ACM Transactions on Mathematical Software},
  volume  = {11},
  number  = {1},
  pages   = {37--57},
  year    = {1985}
}

@article{Orendurff2008,
  author    = {Orendurff, Michael S. and Schoen, Jason A. and Bernatz, Greta C. and Segal, Ava D. and Klute, Glenn K.},
  title     = {How Humans Walk: Bout Duration, Steps per Bout, and Rest Duration},
  journal   = {Journal of Rehabilitation Research and Development},
  volume    = {45},
  number    = {7},
  pages     = {1077--1090},
  year      = {2008},
  doi       = {10.1682/JRRD.2007.11.0197}
}

@article{MoulaeeConradsson2025,
  author    = {Moulaee Conradsson, David and Aktar, Burcin and Bezuidenhout, Lucian},
  title     = {Characterizing Physical Activity Bouts in People with Stroke with Different Ambulation Statuses},
  journal   = {PLoS One},
  volume    = {20},
  number    = {5},
  pages     = {e0307625},
  year      = {2025},
  doi       = {10.1371/journal.pone.0307625}
}

@inproceedings{Pellegrini2020,
  author    = {Pellegrini, Lorenzo and Graffieti, Gabriele and Lomonaco, Vincenzo and Maltoni, Davide},
  title     = {Latent Replay for Real-Time Continual Learning},
  booktitle = {Proceedings of the IEEE/RSJ International Conference on Intelligent Robots and Systems (IROS)},
  pages     = {10203--10209},
  year      = {2020},
  doi       = {10.1109/IROS45743.2020.9341460}
}

@inproceedings{Hayes2020,
  author    = {Hayes, Tyler L. and Kafle, Kushal and Shrestha, Robik and Acharya, Manoj and Kanan, Christopher},
  title     = {{REMIND} Your Neural Network to Prevent Catastrophic Forgetting},
  booktitle = {Computer Vision -- ECCV 2020},
  series    = {Lecture Notes in Computer Science},
  pages     = {466--483},
  year      = {2020},
  doi       = {10.1007/978-3-030-58598-3_28}
}

@article{Yuan2024,
  author    = {Yuan, Hang and Chan, Shing and Creagh, Andrew P. and Tong, Catherine and Acquah, Aidan and Clifton, David A. and Doherty, Aiden},
  title     = {Self-supervised learning for human activity recognition using 700,000 person-days of wearable data},
  journal   = {npj Digital Medicine},
  volume    = {7},
  pages     = {91},
  year      = {2024},
  doi       = {10.1038/s41746-024-01062-3}
}

@inproceedings{schaul2016prioritized,
  title     = {Prioritized Experience Replay},
  author    = {Schaul, Tom and Quan, John and Antonoglou, Ioannis and Silver, David},
  booktitle = {International Conference on Learning Representations (ICLR)},
  year      = {2016}
}

\end{document}